\documentclass[sigconf, nonacm]{acmart}

\renewcommand\footnotetextcopyrightpermission[1]{}
\usepackage{booktabs}
\usepackage{xcolor}
\usepackage{colortbl}
\usepackage{makecell}
\usepackage{array}
\usepackage{placeins}
\usepackage{listings}

\definecolor{ebblue}{HTML}{2171B5}
\definecolor{ebbluepale}{HTML}{C6DBEF}
\definecolor{eborange}{HTML}{E6550D}
\definecolor{eborangepale}{HTML}{FDBE85}
\definecolor{ebcoral}{HTML}{E34A33}
\definecolor{ebcontamination}{HTML}{FFF3CD}

\hypersetup{
  colorlinks=true,
  linkcolor=ebblue,
  citecolor=ebblue,
  urlcolor=ebblue,
}

\begin{document}

\title{Epistemic Blinding: An Inference-Time Protocol for Auditing Prior Contamination in LLM-Assisted Analysis}

\author{Michael F. Cuccarese, PhD}
\affiliation{%
  \country{}}
\email{}

\begin{abstract}
This paper presents epistemic blinding in the context of an agentic system that uses large language models to reason across multiple biological datasets for drug target prioritization. During development, it became apparent that LLM outputs silently blend data-driven inference with memorized priors about named entities---and the blend is invisible: there is no way to determine, from a single output, how much came from the data on the page and how much came from the model's training memory. \textit{Epistemic blinding} is a simple inference-time protocol that replaces entity identifiers with anonymous codes before prompting, then compares outputs against an unblinded control. The protocol does not make LLM reasoning deterministic, but it restores one critical axis of auditability: measuring how much of an output came from the supplied data versus the model's parametric knowledge. The complete target identification system is described---including LLM-guided evolutionary optimization of scoring functions and blinded agentic reasoning for target rationalization---with demonstration that both stages operate without access to entity identity. In oncology drug target prioritization across four cancer types, blinding changes 16\% of top-20 predictions while preserving identical recovery of validated targets. The contamination problem is shown to generalize beyond biology: in S\&P~500 equity screening, brand-recognition bias reshapes 30--40\% of top-20 rankings across five random seeds. To lower the barrier to adoption, the protocol is released as an open-source tool and as a Claude Code skill that enables one-command epistemic blinding within agentic workflows. The claim is not that blinded analysis produces better results, but that without blinding, there is no way to know to what degree the agent is adhering to the analytical process the researcher designed.
\end{abstract}

\maketitle

%% ====================================================================
%% 1. INTRODUCTION
%% ====================================================================
\section{Introduction}

Consider a concrete example. An LLM is given a table of 100 genes resulting from a target discovery model in colorectal cancer and asked to rank the top~20 drug targets. When the gene names are visible, the model ranks KRAS at position~\#1 and justifies the choice by citing ``proven therapeutic tractability via covalent RAS inhibitors.'' When the identical features are presented with anonymous labels, the gene corresponding to KRAS---now labeled Gene\_088---is ranked~\#5, below genes with higher mutation frequencies and stronger convergence signals, but still clearly a target of interest. The phrase ``proven therapeutic tractability via covalent RAS inhibitors'' does not appear anywhere in the provided data. It comes from the model's training corpus.

The past decade has produced a rapidly growing set of publicly accessible, uniform data layers for drug discovery---genome-wide association studies, CRISPR essentiality screens, protein foundation model embeddings, single-cell transcriptomics, and more. The opportunity to deploy modern machine learning methods across these modalities to discover novel therapeutic targets is well understood~\cite{hoadley2018,boiko2023}. Large language models offer two specific advantages in this context. First, they can reason across heterogeneous data layers, integrating signals that span different biological modalities. Second, for well-defined optimization objectives, LLMs can serve as mutation operators in evolutionary frameworks, rapidly improving scoring functions without manual feature engineering.

However, both of these functions are contaminated by a shared failure mode: \textit{parametric knowledge contamination}. When an LLM encounters a named entity during reasoning or code generation, it activates everything it memorized about that entity during pretraining. For drug target prioritization, this means the model has absorbed strong priors about which genes are ``important'' in which cancers---and these priors silently influence outputs that are supposed to be data-driven. The same problem applies whenever LLMs reason over named entities with uneven representation in the training corpus: equities, legal cases, research papers, candidate catalysts, job applicants.

The solution presented here---\textit{epistemic blinding}---draws on the same principle as blinding in clinical trials: prevent the analyst from accessing information that could bias the analysis. Here, the analyst is an LLM, and the blinding is achieved through string replacement at the prompt level.

\paragraph{Outcome impact as distinct from analytical purity.} The natural question is whether blinded or unblinded analysis produces ``better'' results. This paper deliberately resists that framing. In some cases, training priors are genuinely informative---a model that knows KRAS is druggable is incorporating real-world knowledge that might genuinely improve a recommendation. In other cases, those same priors mask data-driven candidates that could represent novel biology. The point of epistemic blinding is not to suppress the model's knowledge. It is to ensure the \textit{influence} of that knowledge visible and measurable---to restore an axis of auditability that is otherwise absent.

For deterministic processes, every step is traceable. A data point can be followed from input through every transformation to output, and the logic verified against intent. LLM-based analysis breaks this: the model's reasoning is a black box in which supplied data and memorized knowledge are mixed without attribution. Epistemic blinding does not make agentic reasoning deterministic. But it brings back something expected of deterministic code: the ability to audit whether the tool is operating on the inputs provided, or on something else entirely.

\paragraph{Contributions.} This work originated in a specific need---building an agentic system for multi-dataset biological reasoning without prior contamination---and generalized into a domain-agnostic protocol. The contributions are: (1)~a formal description of epistemic blinding, including treatment of cross-dataset consistency and subtle leak sources; (2)~an open-source, config-driven implementation that works with any LLM; (3)~a complete target identification system in which epistemic blinding is applied to both evolutionary optimization and agentic reasoning; (4)~a controlled experiment showing systematic fame bias across four cancer types; and (5)~a demonstration in S\&P~500 equity screening showing that the same contamination problem---and the same fix---transfers to an unrelated domain.

%% ====================================================================
%% 2. RELATED WORK
%% ====================================================================
\section{Related Work}

\paragraph{Data contamination in LLM evaluation.} The problem of training data leaking into benchmark evaluations is well-studied~\cite{oren2024,golchin2024,sainz2023}. The present work addresses a related but distinct problem: the model's \textit{general knowledge} contaminating \textit{domain-specific reasoning}, not memorized test sets inflating metrics.

\paragraph{Entity bias in language models.} Wang et al.~\cite{wang2023} demonstrated entity-specific biases in factual question answering and proposed causal interventions that perturb entity names. The present work extends this from factual QA to scientific and financial analysis, where the consequences extend beyond accuracy to resource allocation and discovery.

\paragraph{Anonymization in multi-agent reasoning.} Choi et al.~\cite{choi2025} showed that anonymizing agent identities in multi-LLM debate reduces conformity bias. The present work extends this principle from social identity to epistemic identity---anonymizing the \textit{objects of reasoning} rather than the \textit{reasoning agents}.

\paragraph{Pretraining leakage in protein language models.} Hermann et al.~\cite{hermann2024} demonstrated that protein language model pretraining creates data leakage that distorts downstream task performance by 11.1\%. Their solution---pretraining-aware data splits---addresses the problem at the data level. Ours addresses it at inference time, requiring no retraining.

\paragraph{LLMs for biological analysis.} Hu et al.~\cite{hu2025} used GPT-4 for gene set function discovery with blinded \textit{human} evaluation of the LLM's outputs. Notably, they did not blind the LLM itself---the model received real gene names and could draw on its training knowledge. The present work completes this design by blinding both sides.

%% ====================================================================
%% 3. METHOD
%% ====================================================================
\section{Method: Epistemic Blinding}

\subsection{Protocol}

Epistemic blinding is an inference-time intervention requiring no model modification. Given one or more datasets containing named entities:

\begin{enumerate}
\item \textbf{Identify entity columns.} Determine which columns contain identifiers that may have uneven representation in the LLM's training corpus (gene symbols, tickers, drug names, team names, author names).

\item \textbf{Build a shared mapping.} For each entity type, collect every distinct value across all datasets. Shuffle deterministically with a stable seed and assign anonymous codes of the form \texttt{\{Prefix\}\_\{number\}} (e.g., Gene\_001, Company\_042). The same entity receives the same code everywhere it appears, preserving cross-dataset reasoning.

\item \textbf{Identify and mitigate subtle leak sources.} Beyond the entity name itself, correlated features may reveal identity. For example, market capitalization and sector together can identify most mega-cap stocks; a 42\% mutation frequency in colorectal cancer effectively identifies KRAS. The protocol distinguishes \textit{obvious} leak sources (the entity name) from \textit{subtle} ones (features that correlate strongly enough with identity to partially unblind the model). The protocol requires the analyst to enumerate potential subtle leaks and decide whether to normalize, bin, or drop them. In the financial case study (Section~5), all features were normalized by sector to reduce structural identifiability.

\item \textbf{Shuffle and render.} Randomize row order with a fixed seed and format the anonymized data as a prompt, preserving column headers, units, and feature descriptions.

\item \textbf{Run A/B comparison.} Send the blinded prompt to an LLM in a fresh session with no prior context. Separately, send the matched unblinded prompt to the same model in an independent session. Compare outputs using set overlap, rank shift, and qualitative justification analysis.

\item \textbf{De-anonymize.} Resolve anonymous codes back to real identifiers using the mapping table.
\end{enumerate}

\begin{figure*}[t]
\centering
\includegraphics[width=\textwidth]{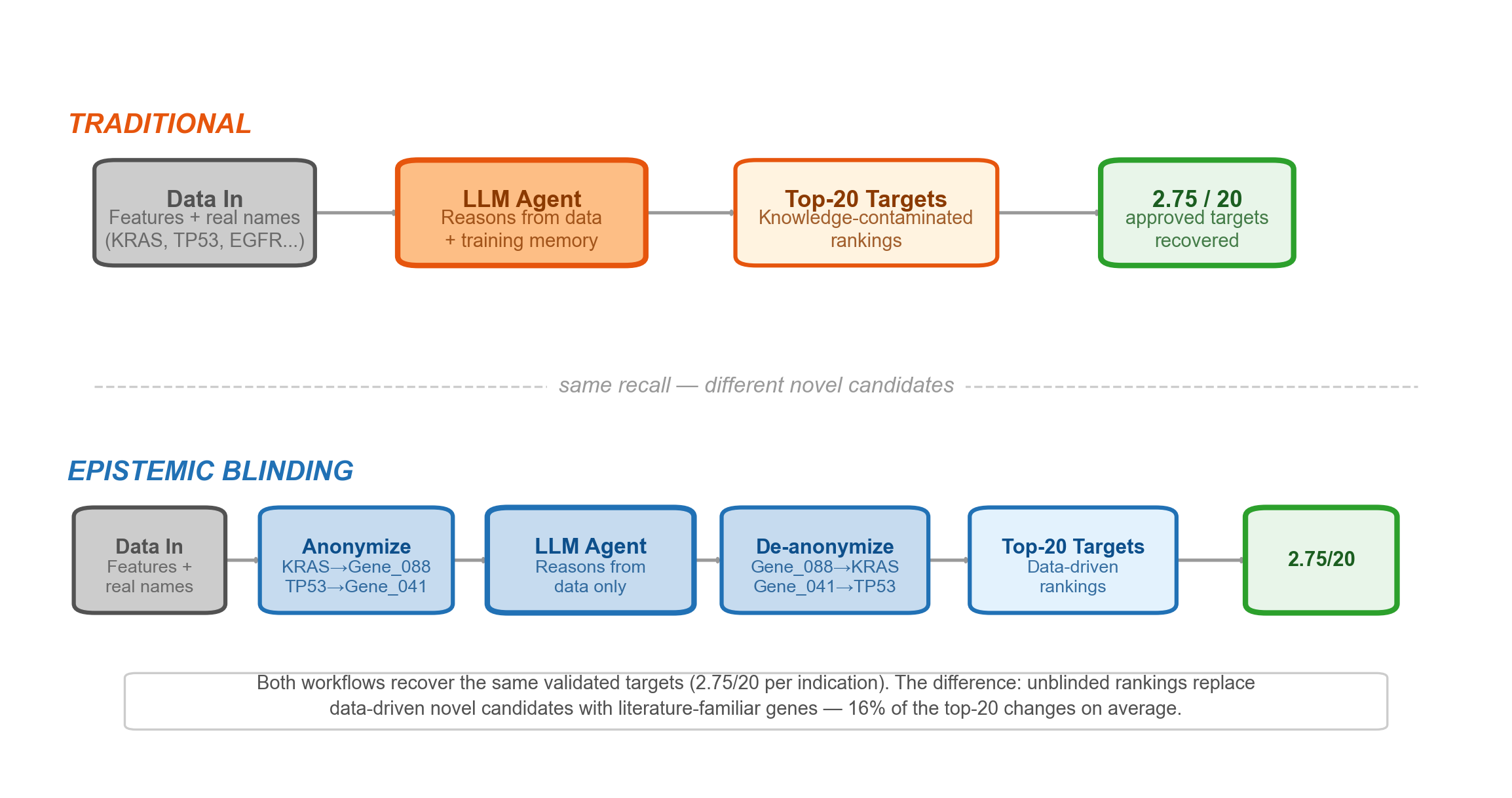}
\caption{Traditional vs.\ epistemic blinding workflow. Both approaches receive identical quantitative features and recover identical numbers of validated drug targets. The critical difference is in which novel candidates are surfaced: the unblinded workflow promotes literature-familiar genes while the blinded workflow ranks purely on feature strength.}
\label{fig:method}
\end{figure*}

The protocol is illustrated in Figure~\ref{fig:method}.

\subsection{What Gets Blinded and What Doesn't}

The goal is to blind information that activates training priors while preserving information the model legitimately needs for reasoning. Entity identifiers and proper nouns with a training-data footprint are blinded. Feature column names, units, numeric values, categorical labels needed for task context, dates, and data source names are preserved. Some categories sit in the middle---disease names, for instance, carry priors but also anchor the analytical task. The protocol supports partial blinding via per-column configuration.

\subsection{Implementation}

The protocol is released as an open-source, config-driven tool\footnote{\url{https://github.com/mcuccarese/epistemic-blinding}} comprising three scripts: \texttt{blind.py} (reads a YAML config specifying datasets, entity columns, and task instructions; produces matched blinded and unblinded prompts plus a mapping file), \texttt{deblind.py} (reverses the mapping in a model response), and \texttt{compare.py} (computes A/B metrics: set overlap, Jaccard index, mean rank delta, Kendall~$\tau$, and optional fame-bias statistics). The tool is model-agnostic: it operates on prompt text, not API internals. It has also been implemented as a Claude Code skill, enabling one-command blinding within an agentic coding workflow.

%% ====================================================================
%% 4. TARGET IDENTIFICATION
%% ====================================================================
\section{Application: Target Identification for Oncology}

Epistemic blinding was developed in the context of a computational target identification system for oncology. The system has two stages: (1)~a deterministic scoring function, evolved by an LLM-guided process, that ranks all human genes by their likelihood of being viable drug targets; and (2)~a blinded LLM reasoning step that rationalizes and refines the top candidates. Both stages operate without access to gene identity.

\subsection{Data Assembly}

The system integrates eight publicly accessible data layers, each selected for a single property: it measures biology without requiring prior knowledge of what the answer should be.

\paragraph{Included sources.} Somatic mutation data from the cBioPortal TCGA PanCancer Atlas~\cite{hoadley2018} ($\sim$1M mutations, 14 cancer studies) at molecular subtype resolution. GWAS associations from Open Targets~\cite{ochoa2023}. Protein structure embeddings from ESM2~\cite{lin2023}, transcriptomic embeddings from Geneformer~\cite{theodoris2023}, and protein function embeddings from ProtT5~\cite{elnaggar2022}. Genetic constraint from gnomAD v4.1 pLI scores. For each gene in a given disease, convergence enrichment was computed as: the overlap between a gene's experimental neighbors (in each embedding space) and the set of significantly mutated genes, normalized by chance expectation.

\paragraph{Deliberate exclusions.} Literature-mined gene--disease relationships, pre-annotated single-cell atlases with canonical marker genes, and expert-curated pathway databases beyond Gene Ontology were excluded from input features. These sources encode prior beliefs of the research community and would re-inject the same publication bias that epistemic blinding is designed to remove. Each included source was accompanied by a context card~\cite{gebru2021} documenting its strengths, known biases, and failure modes.

\paragraph{Feature set.} Each gene is characterized by seven numerical features per disease: somatic mutation frequency, binary mutation significance, pLI (loss-of-function intolerance), convergence enrichment from each of the three foundation models, and total neighbor count. These features are purely quantitative---a gene's enrichment of 15.4 carries the same information whether the gene is labeled SCN1A or Gene\_016.

\subsection{Evolutionary Optimization of Scoring Functions}

Before applying epistemic blinding to LLM \textit{reasoning}, it was applied structurally to LLM-guided \textit{optimization}. ShinkaEvolve, an evolutionary framework in which an LLM (Claude Sonnet) serves as the mutation operator, to evolve scoring functions that rank genes by their likelihood of being validated drug targets.

\paragraph{Blinding by design.} The scoring functions operate exclusively on a 22-dimensional numerical feature vector---indexed by position (feature~0 through feature~21), never by gene symbol. The LLM's role is to propose improved Python code that transforms this numerical array into a scalar score. At no point during evolution does the LLM see gene names, disease names, or any identifier that could activate training priors about specific targets. Gene symbols are mapped back from Ensembl IDs only after ranking, for human reporting.

\paragraph{Fitness function.} For each of 18 diseases (15 oncology at molecular subtype resolution, 3 non-oncology from GWAS), score all $\sim$20,000 human genes, rank by score, and compute the mean percentile of FDA-approved drug targets among non-targets. Secondary metric: strict hit rate (fraction of diseases with $\geq$1 approved target in the top~20).

\paragraph{Evolution trace.} The optimization ran for 13 generations, producing 53 candidate scoring functions. Key transitions:

\textbf{Generation~0 (baseline).} Naive max-of-3 enrichment: \texttt{score = max(esm2, geneformer, prottrans)}. Ignores all disease context and gene properties. Fitness: 68\%.

\textbf{Generation~3 (geometric mean).} Geometric mean of enrichments naturally rewards multi-modality convergence---a gene high in all three modalities scores higher than one extreme in a single modality. Fitness:~77\%.
\begin{lstlisting}
score = (esm2_enrich * gf_enrich * pt_enrich) ^ (1/3)
\end{lstlisting}
\vspace{-0.5em}

\textbf{Generation~6 (sigmoid gates).} Enrichment ratios are unreliable when a modality has fewer than $\sim$5 neighbors. Per-modality sigmoid gates suppress spurious high enrichment from low-powered modalities. Fitness: 82\%.
\begin{lstlisting}
gate = 1 / (1 + exp(-0.5 * (n_neighbors - 5)))
gated_enrichment = enrichment * gate
\end{lstlisting}
\vspace{-0.5em}

\textbf{Generation~9 (disease routing).} Oncology (somatic mutations) and non-oncology (GWAS) require fundamentally different evidence streams. The model created parallel scoring branches with disease-type-specific gates.

\textbf{Generation~12 (evidence-first hierarchy).} The major conceptual breakthrough: drug targets cluster into two natural tiers. \textbf{Tier~1} genes have direct genetic evidence (significantly mutated or GWAS-hit); they receive a large base score (8.0) with convergence and constraint as tiebreakers. \textbf{Tier~2} genes sit in the pathway neighborhood of Tier~1 genes; they receive a lower base score (3.0) gated by convergence alone. This captures both directly mutated targets (BRAF, KRAS) and their pathway neighbors (MEK, RAF1)---a distinction the LLM discovered without ever seeing a gene name. Strict hit rate: 10/18 diseases with at least one approved target in the top~20.
\begin{lstlisting}
tier1 = is_mutated * mutation_gate
score = tier1 * 8.0              # evidence-first base
      + tier1 * convergence * 2.0  # convergence tiebreaker
      + tier1 * pLI * 1.5          # constraint tiebreaker
      + (1 - tier1) * conv * 0.3   # tier 2 baseline
\end{lstlisting}
\vspace{-0.5em}

The entire trajectory---from naive enrichment to a hierarchical, disease-context-aware scoring system---was discovered by an LLM operating on blinded numerical features. No gene name, disease name, or drug name was visible at any point during optimization.

\FloatBarrier

\subsection{Blinded LLM Reasoning for Target Prioritization}

With the evolved scoring function providing a ranked candidate list, epistemic blinding was applied to the second stage: LLM reasoning over the top candidates. For each of four oncology indications spanning a range of feature signal strength, the top~100 candidate genes were selected using the deterministic scoring function (Section~4.2)---no gene names or identities enter the selection---and shuffled to remove positional cues.

Matched blinded and unblinded prompts were presented to Claude (Anthropic) in fresh sessions with instructions to rank the top~20 targets based exclusively on the provided features.

\begin{table}[h]
\centering
\small
\begin{tabular}{@{}lccc@{}}
\toprule
\textbf{Indication} & \textbf{Sig.\ genes} & \textbf{Targets} & \textbf{Signal} \\
\midrule
Acute myeloid leukemia & 30 & 6 & Strong \\
Pancreatic adenocarcinoma & 32 & 1 & Concentrated \\
Chromosomally unstable CRC & 300 & 1 & Moderate \\
IDH-wildtype glioblastoma & 110 & 3 & Weak \\
\bottomrule
\end{tabular}
\caption{Indications tested, spanning a range of feature signal strength.}
\label{tab:indications}
\end{table}

\subsection{Results}

\paragraph{Aggregate.} Average set overlap between blinded and unblinded top-20 lists was 84\%, meaning blinding changes 16\% of top-20 predictions. Critically, target recovery was identical in both conditions across all four indications (average 2.75 approved targets per indication). Blinding does not degrade recovery of validated biology---it changes which \textit{novel} candidates are nominated.

\begin{table}[h]
\centering
\small
\begin{tabular}{@{}lcccc@{}}
\toprule
& \textbf{Overlap} & \textbf{Only Blind} & \textbf{Only Unblind} & \textbf{Targets} \\
\midrule
AML & 90\% & 2 & 2 & 6/6 \\
Pancreatic & 80\% & 4 & 4 & 1/1 \\
CRC & 90\% & 2 & 2 & 1/1 \\
GBM & 75\% & 5 & 5 & 3/3 \\
\midrule
\textbf{Average} & \textbf{84\%} & 3.25 & 3.25 & 2.75 \\
\bottomrule
\end{tabular}
\caption{Set overlap between blinded and unblinded top-20 lists. Target recovery is identical in both conditions.}
\label{tab:overlap}
\end{table}

\begin{figure}[h]
\centering
\includegraphics[width=\columnwidth]{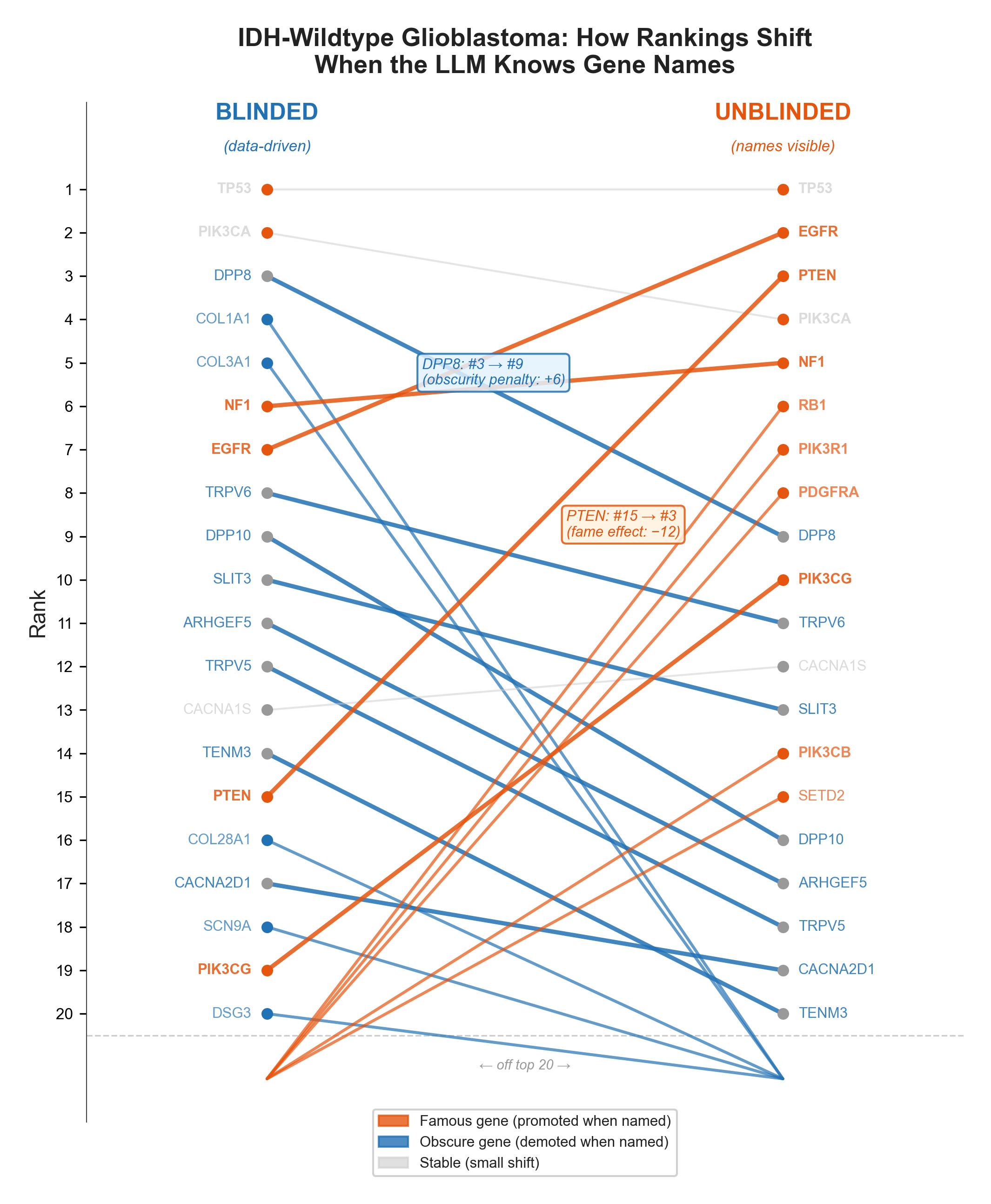}
\caption{Rank-shift slope chart for IDH-wildtype glioblastoma. Lines connect the same gene's rank in the blinded (left) and unblinded (right) conditions. Orange: famous genes promoted when named. Blue: obscure genes demoted when named.}
\label{fig:slope}
\end{figure}

\paragraph{Fame bias.} The direction of change is systematic (Figure~\ref{fig:slope}). Well-known genes are promoted when named, and obscure genes with strong features are demoted:

\begin{table}[h]
\centering
\footnotesize
\setlength{\tabcolsep}{3pt}
\begin{tabular}{@{}llccc>{\raggedright\arraybackslash}p{2.8cm}@{}}
\toprule
\textbf{Gene} & \textbf{Indication} & \textbf{Blinded} & \textbf{Unblinded} & \textbf{Shift} & \textbf{What the model cited} \\
\midrule
PTEN & GBM & \#15 & \#3 & $-$12 & ``the defining PI3K/AKT pathway tumor suppressor'' \\
RNF43 & Pancreatic & \#20 & \#6 & $-$14 & ``Wnt-pathway convergence'' \\
KRAS & CRC & \#5 & \#1 & $-$4 & ``proven therapeutic tractability via covalent RAS inhibitors'' \\
DPP8 & GBM & \#3 & \#9 & $+$6 & Highest triple enrichment in dataset (ESM2=60.9, GF=30.5, PT=16.6) \\
SCN1A & CRC & \#2 & \#11 & $+$9 & Best triple-modality convergence \\
KCNH7 & CRC & \#6 & \#17 & $+$11 & Triple convergence, pLI=0.91, 1252 neighbors \\
\bottomrule
\end{tabular}
\caption{Fame bias: rank shifts between blinded and unblinded conditions. Negative shift means the gene was promoted when named.}
\label{tab:famebias}
\end{table}

DPP8 had the strongest convergence signal in the entire GBM gene set and ranked \#3 when blinded. When named, it dropped to \#9---displaced by genes the model recognized as established GBM biology (RB1, PIK3R1, PDGFRA) despite their weaker features.

\paragraph{Contamination scales with feature ambiguity.} The relationship between signal strength and contamination is consistent and predictive. AML (strong signal, few drivers) shows 90\% overlap. GBM (weak signal, many genes at similar low frequencies) shows 75\% overlap and produces the largest single shift (PTEN, $-$12 ranks). When features clearly differentiate candidates, the model follows the data. When features are ambiguous, the model fills the gap with training knowledge.

\subsection{The Smoking Gun: Justifications Reveal the Mechanism}

The most direct evidence of contamination comes from the model's own justifications (Table~\ref{tab:smokinggun}). When blinded:

\begin{quote}
Gene\_088: ``Very high mutation frequency (42.1\%) with full constraint and strong structural enrichment (ESM2=22.3), pointing to a recurrent constrained driver.''
\end{quote}

When unblinded:

\begin{quote}
KRAS: ``Highest-impact druggable oncogene\ldots\ with \textbf{proven therapeutic tractability via covalent RAS inhibitors} and the strongest combined mutation frequency plus structural convergence in this set.''
\end{quote}

The phrase ``proven therapeutic tractability via covalent RAS inhibitors'' is not derivable from any provided feature. It is parametric knowledge, injected into what was requested to be data-driven analysis. This is not an edge case---it is the mechanism operating in plain sight.

\begin{table}[h]
\centering
\footnotesize
\setlength{\tabcolsep}{3pt}
\begin{tabular}{@{}p{1.2cm}p{3cm}p{3cm}@{}}
\toprule
& \cellcolor{ebbluepale}\textcolor{ebblue}{\textbf{Blinded (\#5)}} & \cellcolor{eborangepale}\textcolor{eborange}{\textbf{Unblinded (\#1)}} \\
\midrule
\textbf{Shared} & \multicolumn{2}{c}{\textit{Both: 38\% mut.\ freq., codon-12 hotspot, strong enrichment}} \\
\addlinespace
\textbf{Extra} & ``GTPase domain mutations suggest gain-of-function'' & \cellcolor{ebcontamination}\textcolor{ebcoral}{\textbf{``Proven tractability via covalent RAS inhibitors (sotorasib, adagrasib)''}} \\
\addlinespace
\textbf{Verdict} & ``Recurrent hotspot with functional convergence'' & ``Defining driver with validated druggability'' \\
\bottomrule
\end{tabular}
\caption{LLM justifications for KRAS / Gene\_088 in CRC. Highlighted text is parametric knowledge not present in any supplied feature.}
\label{tab:smokinggun}
\end{table}

\FloatBarrier

%% ====================================================================
%% 5. S&P 500
%% ====================================================================
\section{Beyond Biology: S\&P 500 Value Screening}

This section may read as quite a topic leap, but it is to illustrate that the contamination problem described above generalizes far beyond disease genetics. Any domain where LLMs reason over named entities with uneven training-corpus representation will exhibit the same phenomenon. To demonstrate this, epistemic blinding was applied to a deliberately different domain: equity analysis.

An LLM was asked to rank the 20 most attractive value investments from $\sim$500 S\&P~500 companies, given four layers of fundamentals: valuation (P/E, P/S, P/B, EV/EBITDA, PEG, FCF yield), growth (revenue, earnings), quality (margins, ROE, ROA, leverage), and shareholder returns (dividend yield). The system prompt was simple---identify undervalued companies and rank strictly by the provided data. Because market capitalization and sector together can identify most mega-cap stocks, all features were presented as normalized values. Ticker symbols were the only entity column blinded.

Five independent seeds (42, 137, 256, 389, 501) were run to estimate variance:

\vspace{-0.5em}
\begin{table}[h]
\centering
\small
\begin{tabular}{@{}lcc@{}}
\toprule
\textbf{Metric} & \textbf{Mean $\pm$ SD} & \textbf{Range} \\
\midrule
Top-20 overlap & 13.0 $\pm$ 1.0 & 12--14 \\
Jaccard index & 0.48 $\pm$ 0.06 & 0.43--0.54 \\
Mean rank delta & 3.1 $\pm$ 1.1 & 1.9--4.4 \\
\bottomrule
\end{tabular}
\caption{S\&P~500 blinded vs.\ unblinded comparison across five seeds.}
\label{tab:sp500}
\end{table}
\vspace{-0.5em}

\begin{figure}[h]
\centering
\includegraphics[width=\columnwidth]{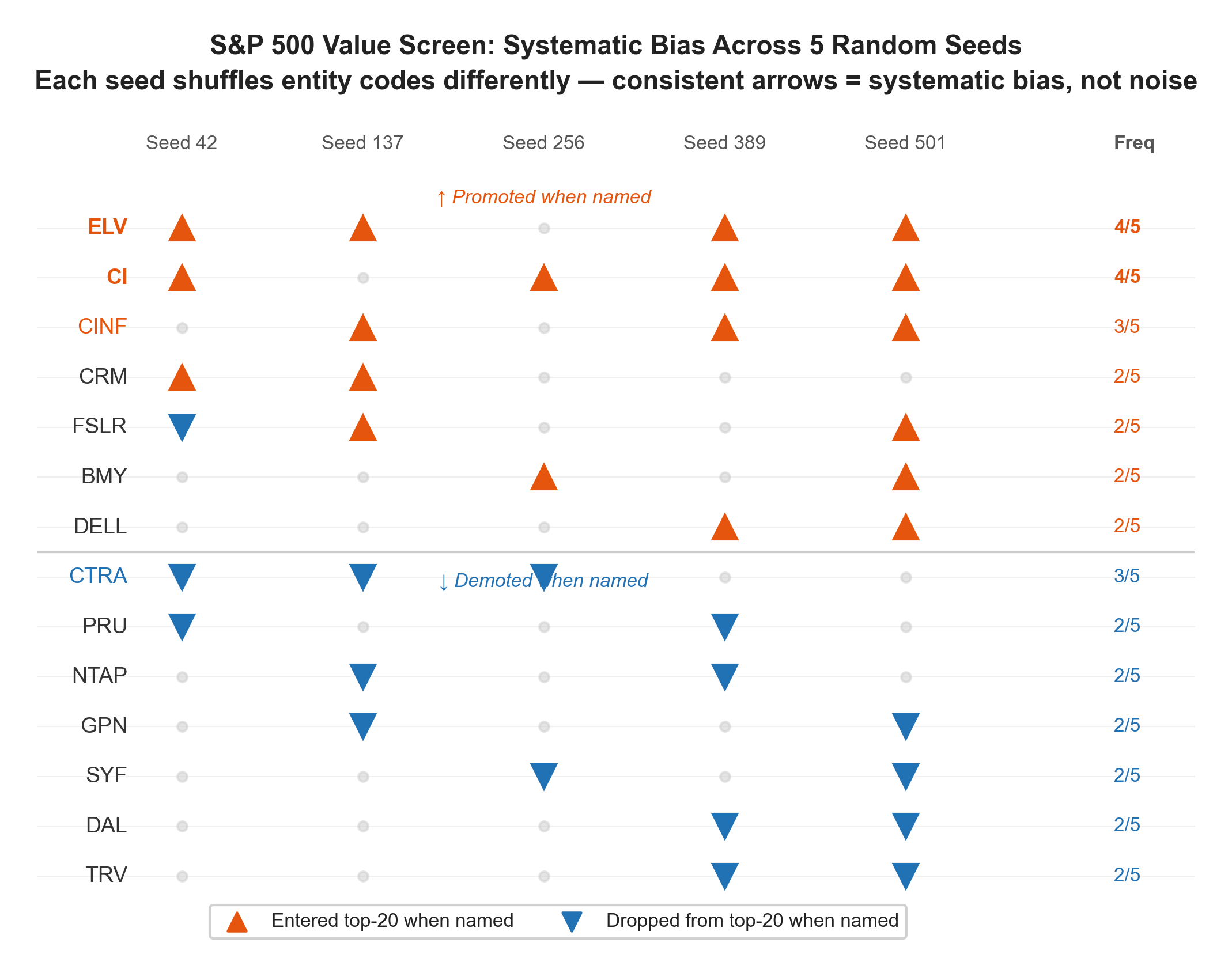}
\caption{S\&P~500 consistency across five random seeds. Consistent arrows indicate systematic bias, not stochastic noise.}
\label{fig:sp500}
\end{figure}

On average, 7 of the top~20 change when tickers are revealed---a 35\% reshaping of the recommendation. The effect is systematic (Figure~\ref{fig:sp500}): tickers like ELV and CI were promoted when unblinded in 4 of 5 runs, while CTRA was demoted in 3 of 5 runs. This is not a claim about equity analysis methodology. It is a demonstration that the same contamination mechanism---LLM priors overriding supplied data---operates identically in a domain with no biological content whatsoever.

\FloatBarrier

\FloatBarrier

%% ====================================================================
%% 6. DISCUSSION
%% ====================================================================
\section{Discussion}

\subsection{Auditability, Not Necessarily Impact}

It is important to be explicit about what epistemic blinding does and does not provide. It provides \textit{auditability}: the ability to measure how much of an LLM's output came from the supplied data versus its training memory. It does not provide \textit{accuracy}: no claim is made that blinded results are more correct than unblinded ones.

In some cases, training priors will be informative. A model that knows KRAS is druggable is incorporating a fact that is true and relevant. In other cases, training priors will be misleading---promoting well-studied genes at the expense of novel candidates with stronger quantitative profiles. The analyst needs to know which situation they are in, and they cannot know that from an unblinded analysis alone.

This framing is analogous to deterministic code. When a traditional program produces an unexpected output, the execution path can be traced to identify exactly where it diverged from expectation. When an LLM-based agent produces an unexpected output, it cannot. Epistemic blinding does not make the agent deterministic---but it restores one specific axis of traceability: did the output come from my data, or from somewhere else?

\subsection{When to Blind}

A practical heuristic: if you would blind a human analyst performing the same task, you should blind the LLM. The protocol is designed for \textit{data-driven inference}---ranking, scoring, prioritizing entities based on supplied features. It is counterproductive for tasks that intentionally leverage the model's knowledge, such as literature synthesis, knowledge retrieval, or hypothesis generation.

Three conditions should hold: (1)~the data contains decision-relevant signal, (2)~the entities have uneven representation in the training corpus, and (3)~someone will act on the output. When all three are true, even a quick A/B comparison provides information the analyst cannot get any other way.

\subsection{What This Does Not Fix}

\paragraph{Sampling bias.} If certain signals are over-represented in data collection, they will be over-represented regardless of blinding. Epistemic blinding operates on the reasoning stage, not the data stage. In the oncology case study presented here, this was addressed by deliberately excluding literature-mined gene--disease associations and other sources where sampling bias would dominate. In domains where data collection itself is biased toward famous entities, blinding the reasoning stage is necessary but not sufficient.

\paragraph{Structural leakage.} Some feature combinations reveal identity even without the name. A company with \$3T market cap and 30\% operating margins in the smartphone sector is Apple regardless of whether the ticker is visible. When structural features are highly identifying, blinding reduces rather than eliminates contamination. Mitigations include normalizing features, binning values, or dropping the identifying column.

\paragraph{Semantically meaningful names.} In some domains, names encode information: IUPAC chemical names encode molecular structure; drug suffixes indicate mechanism (-mab for monoclonal antibodies). Blinding these names destroys legitimate signal. The protocol should be applied to identity, not to nomenclature that carries functional meaning.

%% ====================================================================
%% 7. LIMITATIONS
%% ====================================================================
\section{Limitations}

\begin{enumerate}
\item \textbf{Single model.} All experiments used Claude (Anthropic). The magnitude of contamination may differ across models with different training data exposure.
\item \textbf{Binary comparison.} Only fully blinded vs.\ fully unblinded conditions were compared. Partial blinding strategies (e.g., blinding only the most famous entities) are unexplored.
\item \textbf{No ground truth for novel candidates.} The discovery frontier can be measured to shift under blinding, but which frontier is ``better'' cannot be determined without prospective experimental validation.
\item \textbf{Run-to-run variance.} LLM outputs are stochastic. While the S\&P~500 experiment used five seeds to estimate variance, the oncology experiment used single runs per indication.
\end{enumerate}

%% ====================================================================
%% 8. PRACTICAL ADOPTION
%% ====================================================================
\section{Practical Adoption: A Claude Code Skill}

A protocol is only useful if practitioners can apply it without friction. To this end, epistemic blinding has been implemented as a Claude Code skill---a modular capability that integrates directly into agentic coding workflows. When a user asks an LLM-based coding assistant to rank, score, or prioritize named entities from a dataset, the skill activates automatically: it reads the data, identifies entity columns, generates a YAML configuration, produces matched blinded and unblinded prompts, and runs the A/B comparison---all within a single conversational turn. The user does not need to write configuration files or run scripts manually.

This matters because the primary barrier to epistemic blinding is not conceptual complexity but operational overhead. The protocol is trivially simple in principle---string replacement---but in practice, analysts working within agentic workflows will not pause to anonymize their data, run a parallel session, and compare outputs unless the tooling makes it effortless. By embedding the protocol as a skill that activates contextually, the aim is to make blinding a default step in LLM-assisted data analysis rather than an afterthought.

%% ====================================================================
%% 9. CONCLUSION
%% ====================================================================
\section{Conclusion}

Revealing entity identifiers to an LLM during data-driven analysis systematically biases its outputs toward well-known entities and away from conclusions supported by the data alone. The intervention to detect this---epistemic blinding---is trivially simple, requires no model modification, and preserves the model's analytical reasoning capabilities.

This work has shown that epistemic blinding can be applied at multiple stages of a computational pipeline: structurally, by designing optimization processes that operate on numerical features rather than named entities; and diagnostically, by comparing blinded and unblinded LLM outputs to quantify prior contamination. In oncology, blinding changed 16\% of top-20 drug target predictions while maintaining identical recovery of validated targets. In equity analysis, blinding changed 35\% of top-20 value rankings. In both domains, the direction of bias was systematic: famous entities were promoted, obscure entities with strong features were demoted.

Epistemic blinding should become standard practice whenever LLMs are used for data-driven analysis of named entities. Not because blinded results are necessarily better---but because without blinding, the analyst has no way to know whether the agent is reasoning from the data provided or from something it memorized long before the conversation began.

\paragraph{Code availability.} The epistemic blinding tool, including configuration schemas, worked examples, and comparison scripts, is available at \url{https://github.com/mcuccarese/epistemic-blinding}.

%% ====================================================================
%% REFERENCES
%% ====================================================================
\bibliographystyle{ACM-Reference-Format}

\begin{thebibliography}{14}

\bibitem{oren2024}
Y.~Oren et~al.
\newblock Proving test set contamination in black box language models.
\newblock In \textit{ICLR}, 2024.

\bibitem{golchin2024}
S.~Golchin and M.~Surdeanu.
\newblock Time travel in {LLMs}: Tracing data contamination in large language models.
\newblock In \textit{ICLR}, 2024.

\bibitem{sainz2023}
O.~Sainz et~al.
\newblock {NLP} evaluation in trouble: On the need to measure {LLM} data contamination for each benchmark.
\newblock In \textit{Findings of EMNLP}, 2023.

\bibitem{wang2023}
F.~Wang et~al.
\newblock A causal view of entity bias in (large) language models.
\newblock In \textit{Findings of EMNLP}, 2023.

\bibitem{choi2025}
H.~K.~Choi et~al.
\newblock When identity skews debate: Anonymization for bias-reduced multi-agent reasoning.
\newblock \textit{arXiv:2510.07517}, 2025.

\bibitem{hermann2024}
L.~Hermann et~al.
\newblock Beware of data leakage from protein {LLM} pretraining.
\newblock In \textit{MLCB, PMLR} 261, 2024.

\bibitem{hu2025}
M.~Hu et~al.
\newblock Evaluation of large language models for discovery of gene set function.
\newblock \textit{Nature Methods}, 22:82--91, 2025.

\bibitem{lin2023}
Z.~Lin et~al.
\newblock Evolutionary-scale prediction of atomic-level protein structure with a language model.
\newblock \textit{Science}, 379:1123--1130, 2023.

\bibitem{theodoris2023}
C.~V.~Theodoris et~al.
\newblock Transfer learning enables predictions in network biology.
\newblock \textit{Nature}, 618:616--624, 2023.

\bibitem{elnaggar2022}
A.~Elnaggar et~al.
\newblock {ProtTrans}: Toward understanding the language of life through self-supervised learning.
\newblock \textit{IEEE TPAMI}, 44:7112--7127, 2022.

\bibitem{gebru2021}
T.~Gebru et~al.
\newblock Datasheets for datasets.
\newblock \textit{Communications of the ACM}, 64:86--92, 2021.

\bibitem{hoadley2018}
K.~A.~Hoadley et~al.
\newblock Cell-of-origin patterns dominate the molecular classification of 10,000 tumors from 33 types of cancer.
\newblock \textit{Cell}, 173:291--304, 2018.

\bibitem{boiko2023}
D.~A.~Boiko et~al.
\newblock Autonomous chemical research with large language models.
\newblock \textit{Nature}, 624:570--578, 2023.

\bibitem{ochoa2023}
D.~Ochoa et~al.
\newblock The next-generation {Open Targets Platform}.
\newblock \textit{Nucleic Acids Research}, 51:D1353--D1359, 2023.

\end{thebibliography}

\end{document}